\documentclass[twocolumn,10pt]{asme2ej}

\usepackage{graphicx} 
\usepackage{hyperref}   
\hypersetup{
	colorlinks=true,
	linkcolor=blue,
	citecolor=blue,
	urlcolor=blue,
}
\usepackage[square,numbers]{natbib}

\usepackage{graphicx}      

\usepackage{amsmath,amssymb,amsfonts}

\usepackage{algorithm}
\usepackage{graphicx}
\usepackage{textcomp}
\usepackage{xcolor}
\usepackage{graphicx}
\usepackage{subfig}

\usepackage{array}
\usepackage{url}
\usepackage{xcolor}
\usepackage{multirow}
\usepackage{verbatim}
\usepackage{hyperref}
\usepackage{bbding}
\usepackage{algorithmic}

\makeatletter
\let\@fnsymbol\@arabic
\makeatother

\title{Eco-Driving Control of Connected and Automated Vehicles using Neural Network based Rollout
\thanks{\textcolor{black}{Paper presented at the 2023 Modeling, Estimation, and Control Conference (MECC 2023), Lake Tahoe, NV, Oct. 2-5. Paper No. MECC2023-46.}} $^,$\thanks{\textcolor{black}{The authors acknowledge the support from the United States Department of Energy, Advanced Research Projects Agency – Energy (ARPA-E) NEXTCAR project (Award Number DE-AR0000794)}}
}

\author{Jacob Paugh
    \affiliation{
	Center for Automotive Research\\
	Department of Mechanical Engineering\\
	The Ohio State University\\
	Columbus, OH 43212, USA\\
    Email: paugh.29@osu.edu
    }	
}
\author{Zhaoxuan Zhu
    \affiliation{
    Senior Engineer\\
	Motional\\
    Boston, Massachusetts, USA

    }	
}
\author{Shobhit Gupta, Marcello Canova, Stephanie Stockar
    \affiliation{
	Center for Automotive Research (CAR)\\
	Department of Mechanical Engineering\\
	The Ohio State University\\
	Columbus, OH 43212, USA\\
    }	
}

\begin{document}

\maketitle    

\begin{abstract}
{\it Connected and autonomous vehicles have the potential to minimize energy consumption by optimizing the vehicle velocity and powertrain dynamics with Vehicle-to-Everything info en route. Existing deterministic and stochastic methods created to solve the eco-driving problem generally suffer from high computational and memory requirements, which makes online implementation challenging.
This work proposes a hierarchical multi-horizon optimization framework implemented via a neural network. The neural network learns a full-route value function to account for the variability in route information and is then used to approximate the terminal cost in a receding horizon optimization. Simulations over real-world routes demonstrate that the proposed approach achieves comparable performance to a stochastic optimization solution obtained via reinforcement learning, while requiring no sophisticated training paradigm and negligible on-board memory. 
}
\end{abstract}

\section{Introduction}
The introduction of connected and automated vehicles (CAVs) has had a significant impact due to its potential to improve efficiency by reducing energy consumption. The increased amount of information the vehicle receives from vehicle-to-infrastructure (V2I), vehicle-to-vehicle (V2V), and  global positioning system (GPS) technology affords CAVs the ability to make context-aware decisions en route for higher travel and fuel efficiency \cite{zhu2022safe}.

An eco-driving problem for CAVs can be formulated as an optimal vehicle speed trajectory that minimizes a given cost functional over a route. Recent work in this field explores the opportunity for leveraging information on surrounding traffic and signalized intersection information, allowing for the harmonization of traffic speed \cite{tajalli2020network}. Recent studies aim to minimize energy consumption by either sequentially optimizing \cite{amini2019sequential}, or co-optimizing the speed and powertrain dynamics \cite{deshpande2022real}. Similarly, the potential of leveraging information from signalized intersections to improve CAVs energy efficiency has been shown in \cite{han2023energy}, while V2V opportunities for efficiency improvements on large-scale traffic scenarios are explored in \cite{hyeon2022potential}.

Different solution methods have been investigated in literature to solve the eco-driving problem. Most of these include the use of Pontryagin's minimum principle (PMP) \cite{uebel2017optimal} and Dynamic Programming (DP) \cite{gupta2019thesis}. Although PMP and DP provide globally optimal solutions for a given route itinerary, it is difficult to solve the problem online due to high computational requirements. Alternatively, the problem is solved hierarchically using DP, namely utilizing multiple horizons \cite{borek2019economic}. This solution method, referred to as the rollout algorithm \cite{deshpande2022real}, involves solving a long-term optimization under nominal conditions first, then solving a short-term optimization to account for variability en route. The long-term optimization is considered a base-heuristic, and it is approximated as the terminal cost for the short-horizon optimization. 

Different methods may be explored to generate the base-heuristic. For example, a full-route deterministic optimization could be performed using DP and stored as multi-dimensional maps \cite{zhu2021gpu}. Alternatively, approaches from the 
field of Machine Learning (ML) which leverage data sets obtained via field experiments or simulation may be applied. For instance, a Safe Model-based Off-policy Reinforcement Learning (SMORL) method was demonstrated to learn the terminal cost offline \cite{zhu2022safe}. \textcolor{black}{In addition, several methods utilize online methods based on Q-learning \cite{lee2020model} and Actor-Critic networks \cite{wegener2021automated} to optimize the base heuristic during simulation. }

Computing a full-route DP solution requires no training online but fails to account for unknown route variability, such as signal phase and timing (SPaT), \textcolor{black}{which refers to the duration and order of a traffic light's states (i.e. red, green and yellow)}.   Moreover, the storing of the pre-computed deterministic value function is memory intensive, hence prohibitive for in-vehicle deployment. Conversely, using ML for offline or online training requires less memory relative to the DP solution, but the extensive training and the feedback process intrinsic to reinforcement learning make this method computationally expensive. \textcolor{black}{More specifically, model based reinforcement learning introduces computational expense given its complexity and while model free reinforcement learning lessens the additional expense, it introduces inaccuracy by relying on a reward function that does not account for system characteristics or physics}. 
\begin{figure}
    \centering
    \includegraphics[width=0.5\textwidth]{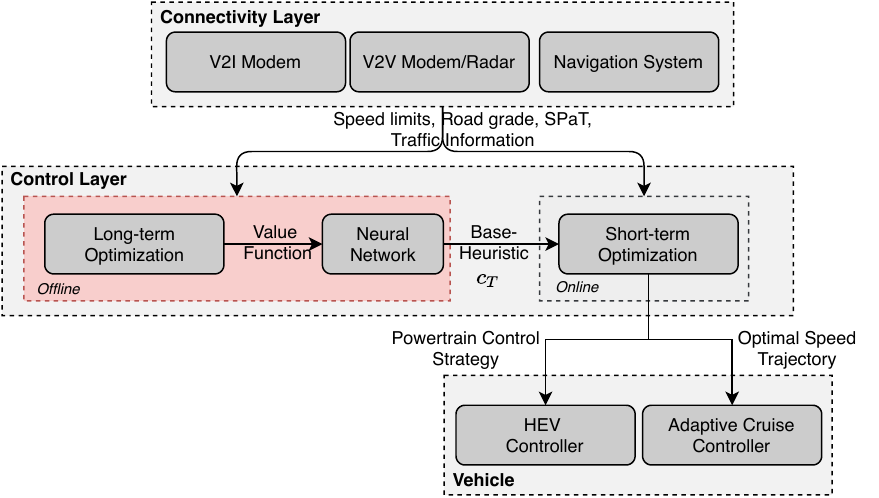}
    \caption{Hierarchical multi-horizon optimization framework}
    \label{fig:control_architecture}
\end{figure}
This work proposes a neural-network (NN) based methodology of extracting the base-heuristic from a pre-computed full-route solution derived from a \textcolor{black}{co-optimization of powertrain dynamics and velocity \cite{deshpande2022real}}. Given a terminal state as an input, the NN approximates the terminal cost for the remainder of the driving mission. Training is performed using the value function computed on different routes and SPaT combinations. The presented approach is illustrated in Fig. \ref{fig:control_architecture} and offers three advantages over the previous methods. \textcolor{black}{First, the offline training of the terminal cost saves significant computation time relative to the SMORL method. Second, the neural network trains on a set of optimal DP solution data \textcolor{black}{based on an accurate physical model} which improves on the accuracy achieved by online reinforcement learning methods. Third, the neural network structure is implemented as a function approximation rather than a static map, significantly decreasing the memory requirements}

\section{Vehicle Dynamics and Powertrain Model}
A forward-looking model of the longitudinal vehicle dynamics and a 48V mild-hybrid powertrain is used in this work for computing energy use \cite{gupta2019thesis}. The powertrain consists of a Belted Starter Generator (BSG) connected to a 1.8L turbocharged gasoline engine. 
The battery is modeled as a zero-th order equivalent circuit to determine the State-of-Charge (SoC). Quasi-static models predict the engine fuel consumption, as well as BSG, torque converter and transmission efficiency. The model was validated with test data collected on a chassis dynamometer for regulatory drive cycles \cite{olin2019reducing}.

\section{Problem Formulation}
\subsection{Full Route Optimization}
Let $s \in [0,N]\subset \mathbb{R}$ denote the discrete distance step, 
 $x_s=[v_s,\xi_s,t_s]\in \mathcal X \subset \mathbb{R}^n$ the state variables comprising of vehicle velocity $v_s$, battery SoC $\xi_s$ and travel time $t_s$,  $u_s=[T_{eng,s},T_{bsg,s}]\in \mathcal U \subset \mathbb{R}^m$ be the control input comprising of the engine $T_{eng,s}$ and BSG torque $T_{bsg,s}$ respectively. The discretized state dynamics is:
 \begin{equation}
    \label{eq: defn_state_dyn}
     x_{s+1} = f(x_s,u_s), \quad s\in [0,N-1] 
\end{equation}
where $x_0$ is the known initial condition for vehicle speed and SoC and \textcolor{black}{$\bar{v}_{s}$ denotes average velocity for a time step}. The equations describing the discrete state dynamics $f(x_s, u_s)$ at distance step $s$ are:
 \begin{subequations}
     \begin{gather}
         v_{s+1}^2=v_s^2+2\Delta d_s \cdot \left(\frac{F_{\mathrm{tr},s}-F_{\mathrm{road},s}(v_s)}{M}\right)\\
         \xi_{s+1}=\xi_s-\frac{\Delta d_s}{\bar{v}_s}\cdot \frac{\bar{I}_{\mathrm{batt},s}}{C_{\mathrm{nom}}}\\
         t_{s+1} = \begin{cases} 
         t_s + t_{\mathrm{RG},s} &, s \in \mathcal{D}_{\mathrm{TL}} \text{ and } \bar{v}_s=0 \\
         t_s + \frac{\Delta d_s}{\bar{v}_{s}} &, s \notin \mathcal{D}_{\mathrm{TL}} 
         \end{cases}
     \end{gather}
 \end{subequations}
 where $F_{\mathrm{tr},s}$ is the tractive force produced by the powertrain \cite{olin2019reducing}. $F_{\mathrm{road},s}$ is the road load resistive force, $M$ is the total vehicle mass, $\bar{I}_{\mathrm{batt},s}$ is the current evaluated over a distance step, $C_{\mathrm{nom}}$ is the nominal battery capacity, $t_s$ is the travel time at a position $s$. Here, $t_{\mathrm{GR},s}$ and $t_{\mathrm{RG},s}$ represent the time remaining in the green and red phase respectively. It is assumed that the positions of all the traffic lights in the route are known a priori from a navigation system, and contained in the set $\mathcal{D}_{\mathrm{TL}}$.

An admissible control map at distance $s$ is denoted by $\mu_s:\mathcal X\rightarrow\mathcal U$, which satisfies the constraint $h(x_s,\mu_s(x))\leq 0$ for all $x_s \in \mathcal X$, where $h: \mathcal{X}\times\mathcal{U}\rightarrow\mathbb{R}^p$,
\begin{subequations}
		\label{eq: constraints_N_horizon}
		\begin{align}
			v_k&\in[v_k^{\mathrm{min}},v_k^{\mathrm{max}}],\\
			\xi_k&\in[\xi_k^{\mathrm{min}},\xi_k^{\mathrm{max}}],\\
			t_k&\in\mathcal{T}_{\mathrm{G},k},\\
			a_k&\in[a^{\mathrm{min}},a^{\mathrm{max}}],\\
			T_{\mathrm{eng},k}&\in[T_{\mathrm{eng}}^{\mathrm{min}}(v_k),T_{\mathrm{eng}}^{\mathrm{max}}(v_k)],\\
			T_{\mathrm{bsg},k}&\in[T_{\mathrm{bsg}}^{\mathrm{min}}(v_k),T_{\mathrm{bsg}}^{\mathrm{max}}(v_k)],
		\end{align}
\end{subequations}
where $v_k^{\{\mathrm{min,max\}}}$, $\xi_k^{\{\mathrm{min,max\}}}$, $a^{\{\mathrm{min,max\}}}$, $T_{\mathrm{eng}}^{\{\mathrm{min,max\}}}$ and $T_{\mathrm{bsg}}^{\{\mathrm{min,max\}}}$ refers to the minimum and maximum limits on speed, SoC, acceleration, engine and BSG torque respectively. $\mathcal{T}_{\mathrm{G},k}$ refers to the green window of the traffic light.
The sequence of admissible control maps denoted by $\mu:=(\mu_s)_{s=0}^{N-1}$ is referred to as the policy of the controller. The set of admissible policies is denoted by $\Gamma$. A running cost (or stage cost) function $c: \mathcal X \times \mathcal U \rightarrow \mathbb{R}$ is introduced, accounting for the trade-off between the fuel consumption and travel time:

\begin{equation} \label{eq: full_route_OCP}
    \resizebox{.9\hsize}{!}{$c(x_k, \mu_{k}(x_{k}))=\left(\gamma \cdot \dot{m}_{\mathrm{f},k}(x_k, \mu_{k}(x_{k}))  +(1-\gamma)\right)\cdot \Delta t_k$}
\end{equation} 

where $\dot{m}_{\mathrm{f},k}(x_k, \mu_{k}(x_{k}))$ is the rate of fuel consumption, $\Delta t_k$ is the travel time over a given distance step and $\gamma$ is the relative weight between fuel and travel time. \textcolor{black}{A comprehensive study on the effect of the selection of $\gamma$ on the optimal solution was performed in \cite{deshpande2022real}.} 

The Optimal Control Problem (OCP) is formulated over a $N$ steps full-route as:
\begin{equation} \label{eq: full_route_OCP}
    \begin{aligned}
    \min_{\mu\in \Gamma} \quad & c_{N}(x_{N}) + \sum_{s = 0}^{N-1} c(x_s,\mu_s(x_s)) \\
     \text{s.t.}\quad & x_{s+1} = f(x_s,\mu_s(x_s)),\\
    & h(x_s,\mu_s(x))\leq 0. 
    \end{aligned}
\end{equation}
where $c_N(x)$ accounts for costs related to the terminal state of the system.

\subsection{Receding Horizon Optimization}
\textcolor{black}{To allow for real-time implementation and account for sources of variability along the route, the optimization is reformulated as a receding horizon OCP (RHOCP) and solved using a Rollout Algorithm, an online suboptimal control technique based on approximation in the value space \cite{deshpande2022real}. Here, the optimal cost-to-go function obtained from solving the full-route optimization is improved upon by solving a one-step look-ahead optimization.} Considering the same discrete dynamic function in Eqn. \eqref{eq: defn_state_dyn} and constraints in Eqn. \eqref{eq: constraints_N_horizon}, at distance $s$, the policy $\bar \mu$ is evaluated by solving the constrained RHOCP:

\begin{equation}
		\label{eq: ed_ocp_N_H_horizon}
		\begin{gathered}
        \begin{aligned}
			\mathcal{J}^*(x_s)= \min_{\left\lbrace \bar \mu_k \right\rbrace_{k=s}^{s+N_H-1}} c_\mathrm{T}(x_{s+N_H})\\+\sum_{k=s}^{s+N_{H}-1}  c(x_k,\bar \mu_{k}(x_{k}))
		\end{aligned}
        \end{gathered}
\end{equation}

where $c_T(x_{s+N_H})$ is the terminal cost, approximated as the base-heuristic from the value function of the full-route optimization in Eqn. \eqref{eq: full_route_OCP}. 

Note that the full-route pre-optimization assumes a fixed SPaT sequence, neglecting the variability that can significantly impact the energy consumption within the RHOCP. To overcome this limitation, a stochastic OCP is defined to minimize the expectation of the cost function over all the possible realizations of SPaT \cite{zhu2022safe}:
\begin{equation}\label{eq: SOCP_formulation}
       \min_{u} \: \mathbb{E} \left[ \sum_{t=0}^\infty \left[\gamma\dot{m}_{\text{fuel},t}+ (1-\gamma) \right]   \Delta t \cdot \mathbb{I}\left[s_t < s_{\text{total}}\right] \right] \\
\end{equation}
where $\mathbb{I}$ is the indicator function, $s_t$ is the distance travelled at time $t$, $s_{\text{total}}$ is the route distance. The constraint set over the $N_H$ horizon remains the same as Eqn. \eqref{eq: constraints_N_horizon}.

The OCP formulated in Eqn. \eqref{eq: SOCP_formulation} is solved using SMORL \cite{zhu2022safe}. The algorithm performs a data-driven approximation of the terminal cost, accounting for the variations in SPaT using a simulator of the vehicle interacting with the environment. The simulated ego-vehicle is assumed to be equipped with GPS and a Dedicated Short Range Communication (DSRC) sensor, providing SPaT data within the communication range. 

When compared against the deterministic rollout solution, which is assumed to be the baseline strategy, SMORL showed  11.0\% less fuel consumption \cite{zhu2022safe}. While the stochastic OCP solved using SMORL allowed the incorporation of a wide range of scenarios, the algorithm requires a large number of online simulations for the training process. In fact, the main limitation of this method is its reliance on an actor-critic network system and a perturbation network supported by both a safe set and a replay buffer \cite{zhu2022safe}. As a direct consequence, this method requires an extremely large amount of memory allocation, which is prohibitive in realistic online implementations.

\subsection{Neural Network based Rollout Algorithm}
Instead of learning the value function using SMORL, a novel rollout algorithm is developed in this paper. This approach requires the full-route optimization in Eqn. \eqref{eq: full_route_OCP} to be run offline for a wide range of simulated routes and SPaT combinations, storing the resulting value functions. A fully connected feed-forward Neural Network (NN) is then trained to approximate the cost-to-go from the terminal cost matrices as a function of an extended state-space. The inputs to the NN are based on an extension of \cite{zhu2022safe} and are summarized in Tab. \ref{tab:InputVector}.

\begin{figure}
    \centering

    \includegraphics[width=0.3\textwidth]{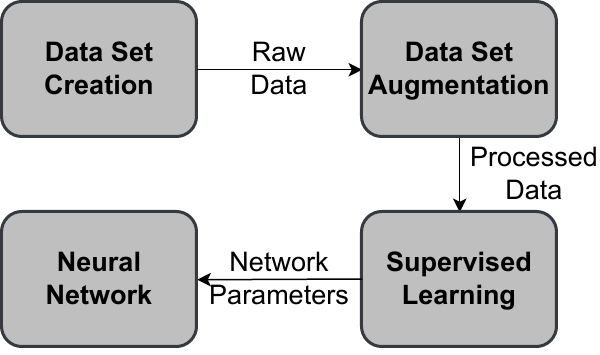}

    \caption{Neural network training flowchart} 
    
    \label{fig:TrainFlowchart}
\end{figure}

The neural networks are trained offline with supervised learning using the processed DP solution as ground truth. This solution benefits the short-term rollout solution as it is trained inexpensively using offline data generation and an offline training process. In addition, the derived network accounts for SPaT, allowing it to function with stochastic elements while requiring significantly less storage space. A summary of the process for producing the neural network is given in Fig. \ref{fig:TrainFlowchart}.

\subsubsection{Data Set Creation and Augmentation}
The 3-state vector is augmented using speed limit and SPaT information under the assumption that each route has information on the placement and the speed limit values used to define its relative difference from vehicle velocity. The route has a known length, therefore distance remaining is an additional state. Then, SPaT is included as the distance to the next traffic light, phase and time sample of the respective phase. The final input vector to the NN for the training takes the form $\bar{X}$ shown in Tab. \ref{tab:InputVector}.

\begin{table}[]
\caption{Input vector form}
\label{tab:InputVector}
\resizebox{\columnwidth}{80pt}{
\begin{tabular}{l|l|l}
              & \textbf{Variable}               & \textbf{Description}                           \\ \hline
 & $SoC\:\in\:\mathbb{R}$                         & Battery SoC                                    \\
              & $V_{veh}\:\in\:\mathbb{R}$                        & Vehicle Velocity                               \\
              & $V_{rlim}\:\in\:\mathbb{R}$                       & Difference of Vehicle Velocity and Speed  \\
              &                                 &Limit at the Current Road Segment                    \\
              & $V'_{rlim}\:\in\:\mathbb{R}$                      & Difference of Vehicle Velocity and    \\
              &                                 &Upcoming Speed Limit                                    \\
 \textbf{$\bar{X}$}             & $d_{tfc}\:\in\:\mathbb{R}$                        & Distance to upcoming traffic light             \\
              & $d'_{lim}\:\in\:\mathbb{R}$                       & Distance to road segment at which the         \\
              &                                 &speed limit changes                                  \\
              & $d_{rem}\:\in\:\mathbb{R}$                        & Remaining distance of the trip                 \\
              & $x_{tfc}\:\in$ & Sampled status of the upcoming traffic   \\
              & ${{\{\mathbb{R}}|-1\le x_{tfc}\le 1\}^6}$                                 &light encoded as six digits                         
\end{tabular}
}
\end{table}

\begin{figure}
  \centering
  \includegraphics[width=0.85\linewidth]{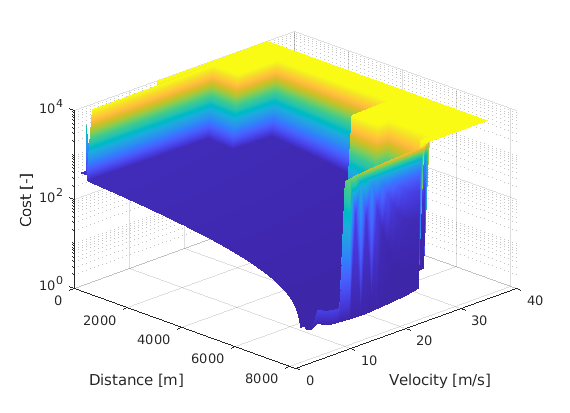}
  \caption{\textcolor{black}{Cost-to-Go graphs as a function of velocity}}
  \label{fig:RawCost_vel}
\end{figure}

\begin{figure}
  \centering
  \includegraphics[width=0.85\linewidth]{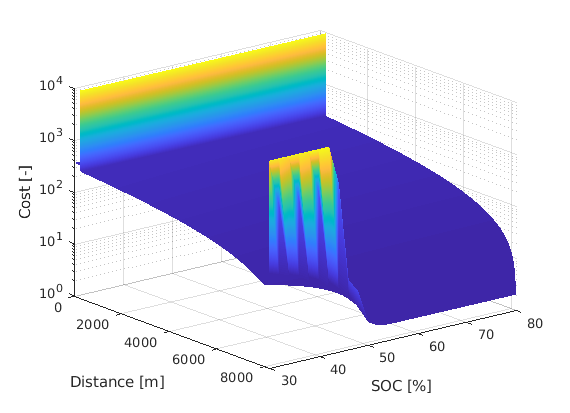}
  \caption{\textcolor{black}{Cost-to-Go graphs as a function of SoC}}
  \label{fig:RawCost_soc}
\end{figure}

\begin{figure}
  \centering
  \includegraphics[width=0.85\linewidth]{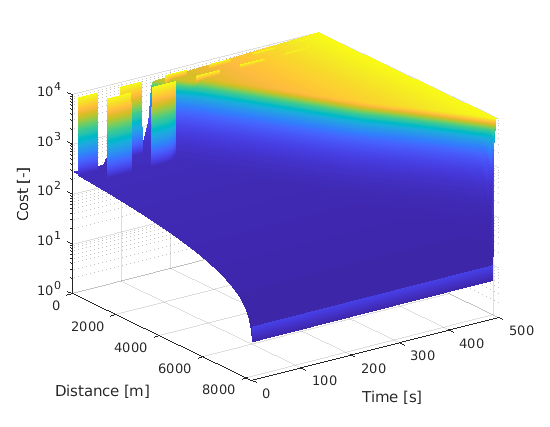}
  \caption{\textcolor{black}{Cost-to-Go graphs as a function of time}}
  \label{fig:RawCost_time}
\end{figure}

The cost-to-go values obtained from DP include high-cost regions that correspond to infeasible operations of the vehicle. Because the objective is to use supervised learning for the approximation of the value function, infeasibilities in the data have been addressed in pre-processing. Specifically, the infeasibilites due to the violation of the limits on the states are shown in Fig. \ref{fig:RawCost_vel}, Fig. \ref{fig:RawCost_soc} and Fig. \ref{fig:RawCost_time}. Infeasibilities in Fig. \ref{fig:RawCost_soc} imposing the recharge constraint were removed because they are applied in rollout. Infeasibilities in Fig. \ref{fig:RawCost_time} for going too slow were removed because they are not applicable in an online simulation scenario. Infeasibilities due to traffic lights were also truncated so that the neural network could account for the obstacle, as in Fig. \ref{fig:RawCost_time}. Finally, infeasibilities from exceeding the speed limit, as shown in Fig. \ref{fig:RawCost_vel}, were removed due to the constraint being applied during rollout.

\begin{algorithm}[]
\begin{algorithmic}[1]

\STATE Initialize neural network, encoder, ADAM loss function, output file for training and test loss\\
\FOR{$n_{iter}$ in N epochs}
    \STATE Initialize loss values\\
    \STATE Randomly sample 100000 points uniformly from the processed DP solution\\
    
    \FOR{$j^{th}$ mini-batch of 500 in the data set}
        \STATE Encode traffic light data\\
        \STATE Perform a forward pass for the input data \\
        \STATE iterate the model weights using back-propagation on prediction error\\
        \STATE Add training loss to running sum\\
    \ENDFOR
    \STATE Average running loss sum for average training loss\\
    \STATE Perform a forward run and calculate loss on test set data using the current network\\
\ENDFOR

 \caption{Neural Network Supervised Learning}
 \label{algorithm: NN}

 \end{algorithmic}
\end{algorithm}

\subsubsection{Training Process} \label{sec: train}
To produce raw data, full-route DP optimization was performed on a set of routes. Once post-processed, this data was used to train the NN whose input layer is the same as the input vector $\bar{X}$, and its output layer is the predicted value function. This neural network is optimized using an offline supervised training process. During training, the data was normalized, scrambled, applied as mini-batches, and the chosen optimization algorithm was ADAM. Using a learning rate of 0.001 and a dropout rate of 30 percent on a two-layer network composed of 500 neurons each, an optimal set of neural network parameters was derived using the algorithm summarized in algorithm \ref{algorithm: NN}. This training process converges by early stopping such that it ceased training once changes in training and testing error became negligible. 

\subsubsection{Neural Network}
The converged NN approximates the terminal cost as a function of the augmented vehicle state input vector $\bar{X}$. Utilizing an offline training process and requiring much less memory to store and use compared to the fully deterministic solution, this new formulation differs from the SMORL formulation by considering stochastic variation in a more computationally efficient manner.

\section{Simulation Results} \label{sec: results}
The trained NN is used as the base-heuristic in the RHOCP Eqn. \eqref{eq: ed_ocp_N_H_horizon} and compared against SMORL. To demonstrate the ability of the NN to predict the terminal cost for a given set of states, an overfit was performed over a training data set \textcolor{black}{ whereby the NN is trained to predict only its training data accurately}. The chosen route was an 8.2 km mixed-urban route with 2 traffic lights\textcolor{black}{, a representative example from the 100-route dataset}.

\begin{figure}
  \centering
  \centerline{\includegraphics[width=\linewidth]{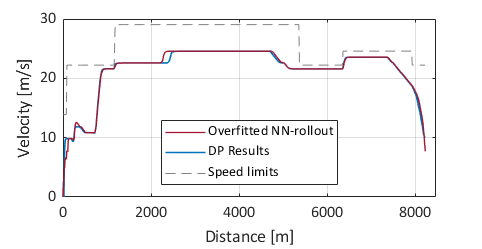}}
\hfill
\caption{Result of neural network overfit to a sample route}
\label{fig:OVFit}
\end{figure}

The trained NN was then integrated in the rollout scheme as shown in Fig. \ref{fig:control_architecture} with a horizon length of 1 km. Figure \ref{fig:OVFit} shows the comparison between the solution of RHOCP Eqn. \eqref{eq: ed_ocp_N_H_horizon} using the deterministic value function and the developed NN as the terminal cost. Results indicate that the velocity trajectories obtained with the two methods are within close proximity, which indirectly confirms that the trained NN is correctly predicting the terminal cost of the RHOCP. 

\begin{figure}
  \centering
  \centerline{\includegraphics[width=\linewidth]{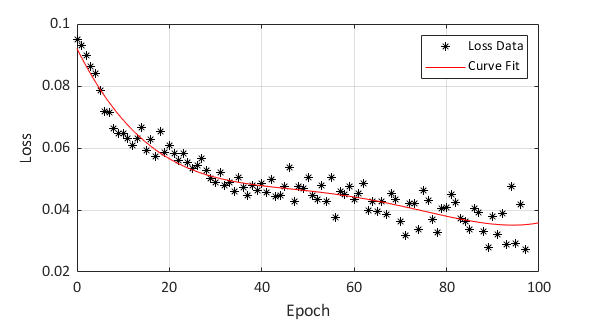}}
\hfill
\caption{Average loss per epoch on the test set}
\label{fig:LossCurve}
\end{figure}

After this initial confirmation, the NN was retrained to generalize and represent several possible routes. A set of 20 real-world routes was chosen with a given SPaT profile generated in SUMO (Simulation of Urban Mobility) \cite{krajzewicz2002sumo}. \textcolor{black}{Each route contains around 2$\cdot10^8$ data points, together composing a set of approximately 4$\cdot10^9$ data points}. For this training process, the data was split into 16 and 4 routes forming the training and test set respectively. The training process detailed in section \ref{sec: train} was followed. Figure \ref{fig:LossCurve} shows the average loss per epoch over the test set, demonstrating convergence during learning.

The resulting NN is integrated with the rollout using a 200m horizon and simulated over five representative routes selected from a set of 100 routes. The routes include three mixed-urban route and two urban routes, summarized in Tab. \ref{tab:RouteChar}.

\begin{table}[]
\caption{Route characteristics}
\label{tab:RouteChar}
\resizebox{\columnwidth}{80pt}{
\begin{tabular}{|l|l|l|l|l|}
\hline
\multirow{2}{*}{Route}               & Length                 & Average Speed          & Traffic            & Stop               \\
                                     & (km)                   & Limit (m/s)            & Lights             & Signs              \\ \hline
Overfit                              & 8.22                   & 25.32                  & 2                  & 2                  \\ \hline
\multirow{2}{*}{Urban Route 1 (UR1)} & \multirow{2}{*}{10.01} & \multirow{2}{*}{18.27} & \multirow{2}{*}{7} & \multirow{2}{*}{2} \\
                                     &                        &                        &                    &                    \\ \hline
\multirow{2}{*}{Urban Route 2 (UR2)} & \multirow{2}{*}{8.41}  & \multirow{2}{*}{19.22} & \multirow{2}{*}{3} & \multirow{2}{*}{2} \\
                                     &                        &                        &                    &                    \\ \hline
\multirow{2}{*}{Mixed-Urban Route 1 (MUR1)} & \multirow{2}{*}{10.83} & \multirow{2}{*}{23.45} & \multirow{2}{*}{5} & \multirow{2}{*}{2} \\
                                     &                        &                        &                    &                    \\ \hline
\multirow{2}{*}{Mixed-Urban Route 2 (MUR2)} & \multirow{2}{*}{7.40}  & \multirow{2}{*}{22.44} & \multirow{2}{*}{6} & \multirow{2}{*}{2} \\
                                     &                        &                        &                    &                    \\ \hline
\multirow{2}{*}{Mixed-Urban Route 3 (MUR3)} & \multirow{2}{*}{7.17}  & \multirow{2}{*}{23.23} & \multirow{2}{*}{7} & \multirow{2}{*}{2} \\
                                     &                        &                        &                    &                    \\ \hline
\end{tabular}
}
\end{table}

\begin{table*}[]
\begin{center}
\caption{Summary statistics of simulated routes}
\label{tab:StatSummaryTable}
\resizebox{450pt}{60pt}{
\begin{tabular}{l|ll|ll|ll|ll|}

\cline{2-9}
                               & \multicolumn{2}{l|}{Fuel Economy (mpg)}                             & \multicolumn{2}{l|}{Time (s)}               & \multicolumn{2}{l|}{Cumulative Cost (-)}        & \multicolumn{2}{l|}{Final SoC (\%)}      \\ \cline{2-9} 
                               & \multicolumn{1}{l|}{SMORL} & NN-Rollout                             & \multicolumn{1}{l|}{SMORL} & NN-Rollout     & \multicolumn{1}{l|}{SMORL}  & NN-Rollout        & \multicolumn{1}{l|}{SMORL}  & NN-Rollout \\ \hline
\multicolumn{1}{|l|}{UR1} & \multicolumn{1}{l|}{44.22} & {43.72 ({\color[HTML]{CB0000} -1.14\%})} & \multicolumn{1}{l|}{752}   & 748 ({\color[HTML]{32CB00} -0.53\%})  & \multicolumn{1}{l|}{594.74} & 594.44 ({\color[HTML]{32CB00} -0.050\%}) & \multicolumn{1}{l|}{0.5006} & 0.5689     \\ \hline
\multicolumn{1}{|l|}{MUR1} & \multicolumn{1}{l|}{42.87} & 45.68 ({\color[HTML]{32CB00} +6.35\%})                        & \multicolumn{1}{l|}{702}   & 712 ({\color[HTML]{CB0000} +1.41\%})  & \multicolumn{1}{l|}{583.96} & 578.43 ({\color[HTML]{32CB00} -0.95\%})  & \multicolumn{1}{l|}{0.5025} & 0.5448     \\ \hline
\multicolumn{1}{|l|}{UR2} & \multicolumn{1}{l|}{43.38} & 46.22 ({\color[HTML]{32CB00} +6.34\%})                        & \multicolumn{1}{l|}{583}   & 596  ({\color[HTML]{CB0000} +2.21\%}) & \multicolumn{1}{l|}{473.29} & 472.11  ({\color[HTML]{32CB00} -0.25\%}) & \multicolumn{1}{l|}{0.5009} & 0.5679     \\ \hline
\multicolumn{1}{|l|}{MUR2} & \multicolumn{1}{l|}{40.08} & 42.52 ({\color[HTML]{32CB00} +5.91\%})                        & \multicolumn{1}{l|}{575}   & 588 ({\color[HTML]{CB0000} +2.24\%})  & \multicolumn{1}{l|}{462.10} & 461.80 ({\color[HTML]{32CB00} -0.065\%}) & \multicolumn{1}{l|}{0.5008} & 0.5648     \\ \hline
\multicolumn{1}{|l|}{MUR3} & \multicolumn{1}{l|}{39.79} & 41.54 ({\color[HTML]{32CB00} +4.30\%})                        & \multicolumn{1}{l|}{532}   & 532 ({\color[HTML]{32CB00} +0\%})     & \multicolumn{1}{l|}{434.31} & 428.91 ({\color[HTML]{32CB00} -1.25\%})  & \multicolumn{1}{l|}{0.5002} & 0.5639     \\ \hline
\end{tabular}
}
\end{center}
\end{table*}

The cumulative performance of the NN-rollout framework is compared against SMORL over the 5 selected routes, as summarized in Tab. \ref{tab:StatSummaryTable}. The results indicate that the NN-rollout method reduces the cumulative cost compared to SMORL for all the routes considered, due to the fact that the NN learned the value function from the global DP solutions. Further, the NN-rollout method improves the fuel economy over SMORL for 4 of the 5 selected routes, with minimal effect on the travel time. 

\begin{figure}
  \centering
  \subfloat[Distance vs velocity trajectory\label{fig:R10V}]{\includegraphics[width=0.45\linewidth]{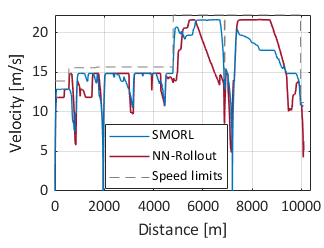}}
\hfill
  \subfloat[Distance vs SoC trajectory\label{fig:R10S}]{\includegraphics[width=0.48\linewidth]{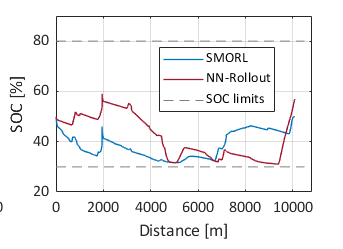}}
  
  \subfloat[Torque vs time trajectory\label{fig:R10T}]{\includegraphics[width=0.45\linewidth]{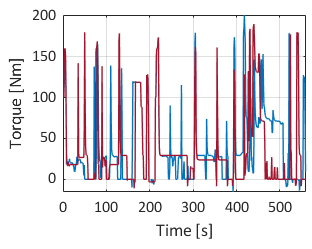}}
\hfill
  \subfloat[Distance vs time trajectory\label{fig:R10D}]{\includegraphics[width=0.48\linewidth]{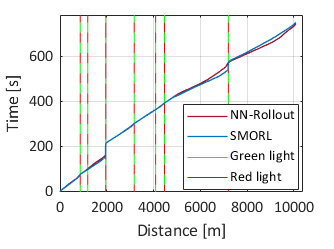}}
  \caption{Urban Route 1 (UR1) trajectory}
  \label{fig:R10NNGraph}
\end{figure}

\begin{figure}
  \centering
  \subfloat[Distance vs velocity trajectory\label{fig:R70V}]{\includegraphics[width=0.45\linewidth]{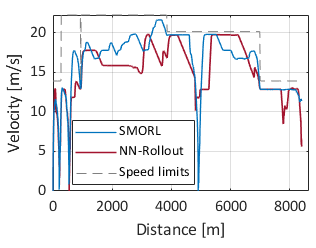}}
\hfill
  \subfloat[Distance vs SoC trajectory\label{fig:R70S}]{\includegraphics[width=0.48\linewidth]{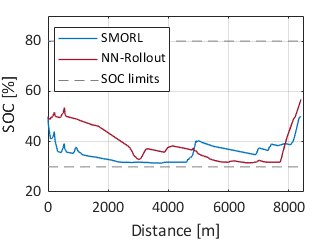}}
  
  \subfloat[Torque vs time trajectory\label{fig:R70T}]{\includegraphics[width=0.45\linewidth]{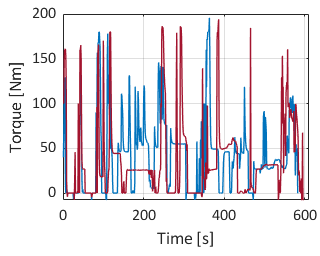}}
\hfill
  \subfloat[Distance vs time trajectory\label{fig:R70D}]{\includegraphics[width=0.48\linewidth]{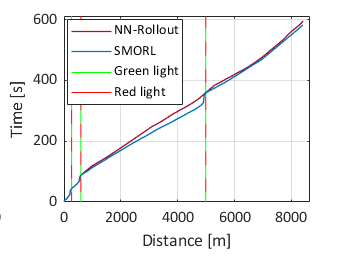}}
  \caption{Urban Route 2 (UR2) trajectory}
  \label{fig:R70NNGraph}
\end{figure}

Figure \ref{fig:R10NNGraph} and Fig. \ref{fig:R70NNGraph} shows the optimal state and control input trajectories, along with the time-space plot for UR1 and UR2, respectively. The state and control input trajectories show that all the constraints on speed limits, battery SoC limits and torque limits are met, which is ensured by the deterministic solution of the short-term RHOCP. A correct prediction of the terminal cost becomes important when approaching intersections, where errors could lead to infeasibilities. To this extent, the results indicate that the vehicle does not violate any red light and stop signs, while passing only when the light is green. 

The velocity trajectory exhibit some noise resulting from a general NN fit, but also exhibits behavior that demonstrates the ability to extrapolate from its training set. 
For example, the vehicle accelerates and then coasts through the speed limit change, which is demonstrated on UR1 between 8000m and 10000m in Fig. \ref{fig:R10V} and on UR2 between 6000m and 7000m in Fig. \ref{fig:R70V}. Constant velocities are held on unobstructed sections of road, which is shown in UR1 between 7000m and 8000m in Fig. \ref{fig:R10V} and UR2 between 5000m and 6000m in Fig. \ref{fig:R70V}. 

The slight change in the velocity trajectory between the NN-based method and SMORL can result in the vehicle experiencing a different SPaT sequence along the route, which is evident from the time-space plot in Fig. \ref{fig:R10NNGraph} and Fig. \ref{fig:R70NNGraph}. For the UR2 case, shown in Fig. \ref{fig:R70V}, the speed profile of the NN-based approach is almost identical to SMORL, except for the NN causing a slower constant velocity between 2000m and 3000m. This is necessary to avoid a red light stop at 5000m, as  shown in Fig. \ref{fig:R70D}, while SMORL encounters a red phase at the same intersection. 

Comparison of the SoC trajectory demonstrates close to charge-sustaining behavior in both cases, with only a slight deviation from the $50\%$ SoC target at the end of the route. This was due to pruning the infeasible raw data shown in Fig. \ref{fig:RawCost_soc}, which discouraged low SoC values at the end of route, creating a preference for slight overshoots. 

Overall, the NN-rollout method can outperform the SMORL approach in approximating the terminal cost. Due to its offline training, the NN-rollout is not only significantly less computationally expensive, but also results in faster simulation times. In addition, compared to the DP (deterministic) solution, the NN-rollout method was able to provide a full approximation of the mapping between terminal state and terminal cost without the need to always store a value function. For the routes considered in this paper, Tab. \ref{tab:DPSize} shows that the NN used 268 kilobytes (kb) of memory compared to 2-5 gigabytes (Gb) needed for the deterministic approach. Therefore, the NN-rollout approach outperforms the existing methods of approximating the terminal cost in a computationally and memory-efficient manner.

\begin{table}[]
\centering
\caption{Memory requirement of approach}
\resizebox{\columnwidth}{25pt}{

\begin{tabular}{l|l|l|l|l|l}
Route                         &UR1 &MUR1 &UR2 &MUR2 &MUR3 \\ \hline
Value Function (GB) & 5.10     & 5.08     & 3.05     & 2.92     & 2.63     \\ \hline
Neural Network (kB) & 269.5    & 269.5    & 269.5    & 269.5    & 269.5   
\end{tabular}
}
\label{tab:DPSize}

\end{table}

\section{Conclusion}
A Neural Network approximation of the terminal cost in a receding horizon optimal control problem is proposed in this paper. The method is compared against a stochastic approach that captures the variability in SPaT. Simulations over five real-world routes show that the proposed NN-rollout outperforms the stochastic method, while reducing the computational and memory requirements. The NN-rollout provides an exhaustive mapping of the terminal cost similar to a full-route DP, but without the significant memory required. 
Additionally, the NN approximation shows similar robustness to variation as the reinforcement learning method, while requiring a completely offline process that is computationally more efficient.

\textcolor{black}{
The current work is focusing on integrating the proposed NN as a terminal cost approximator of the rollout algorithm for Eco-Driving presented in \cite{deshpande2022real}. This will enable the integration in vehicle and experimental verification of the proposed strategy.
}
Future work will focus on extending the NN framework to include uncertainties due to variations in traffic density.

\bibliographystyle{asmems4}
\bibliography{references}             
\end{document}